\documentclass[letterpaper, 10 pt, conference]{ieeeconf}  
\usepackage{amsmath}
\usepackage{graphicx}
\usepackage[font=small]{caption}
\usepackage{subcaption}
\usepackage{comment}
\usepackage{multirow}

\IEEEoverridecommandlockouts                              

\title{\LARGE \bf 
CoboGuider: Haptic Potential Fields for Safe Human-Robot Interaction 
}
\author{Viktor Rakhmatulin$^{1}$, Miguel Altamirano Cabrera$^{2}$, Fikre Hagos$^{2}$, Oleg Sautenkov$^{2}$, Jonathan Tirado$^{2}$, \\ Ighor Uzhinsky$^{1}$, and Dzmitry Tsetserukou$^{2}$
\thanks{$^{2}$ The authors are with the Center for Design, Manufacturing and Materials, and $^{1}$ Space Center of Skolkovo Institute of Science and Technology (Skoltech), 121205 Bolshoy Boulevard 30, bld. 1, Moscow, Russia. {\tt\small \{viktor.rakhmatulin, miguel.altamirano, fikre.hagos, oleg.sautenkov, jonathan.tirado, I.Uzhinsky, D.Tsetserukou\}@skoltech.ru\ }}
}

\begin{document}
    
\maketitle
\thispagestyle{empty}
\pagestyle{empty}

\begin{abstract}
 Modern industry still relies on manual manufacturing operations and safe human-robot interaction is of great interest nowadays. Speed and Separation Monitoring (SSM) allows close and efficient collaborative scenarios by maintaining a protective separation distance during robot operation. The paper focuses on a novel approach to strengthen the SSM safety requirements by introducing haptic feedback to a robotic cell worker. Tactile stimuli provide early warning of dangerous movements and proximity to the robot, based on the human reaction time and instantaneous velocities of robot and operator. A preliminary experiment was performed to identify the reaction time of participants when they are exposed to tactile stimuli in a collaborative environment with controlled conditions. In a second experiment, we evaluated our approach into a study case where human worker and cobot performed collaborative planetary gear assembly. Results show that the applied approach increased the average minimum distance between the robot's end-effector and hand by 44\% compared to the operator relying only on the visual feedback. Moreover, the participants without the haptic support have failed several times to maintain the protective separation distance.

\end{abstract}

\section{Introduction}

Industrial robots have made the manufacturing process cheaper and faster. However, there is still a wide variety of tasks that are difficult to automate, or the automation is not economically justified \cite{Ajoudani2018}. Collaborative robots (cobots) have been specially designed to ensure collaboration between human workers and robots but above all the human safety. A well-known case is the KUKA LBR iiwa robot, which is designed for direct and safe human-robot collaboration. The emergency stop is executed upon the collision detection by the embedded joint torque sensors, which may pin the human worker against a static object by a braked robot arm \cite{Zhang2015}. Moreover, cobots with high payload or sharp tools still represent a dangerously high risk to humans during physical contact. Therefore, to reduce the possibility of human harm, collision-free human-robot interaction is highly desirable. The standards and approaches for collaborative tasks and safe human-robot interaction are actively developing. \par 

\begin{figure}[h!]
  \vspace{0.3cm}
  \centering
  \includegraphics[width=1.0\linewidth]{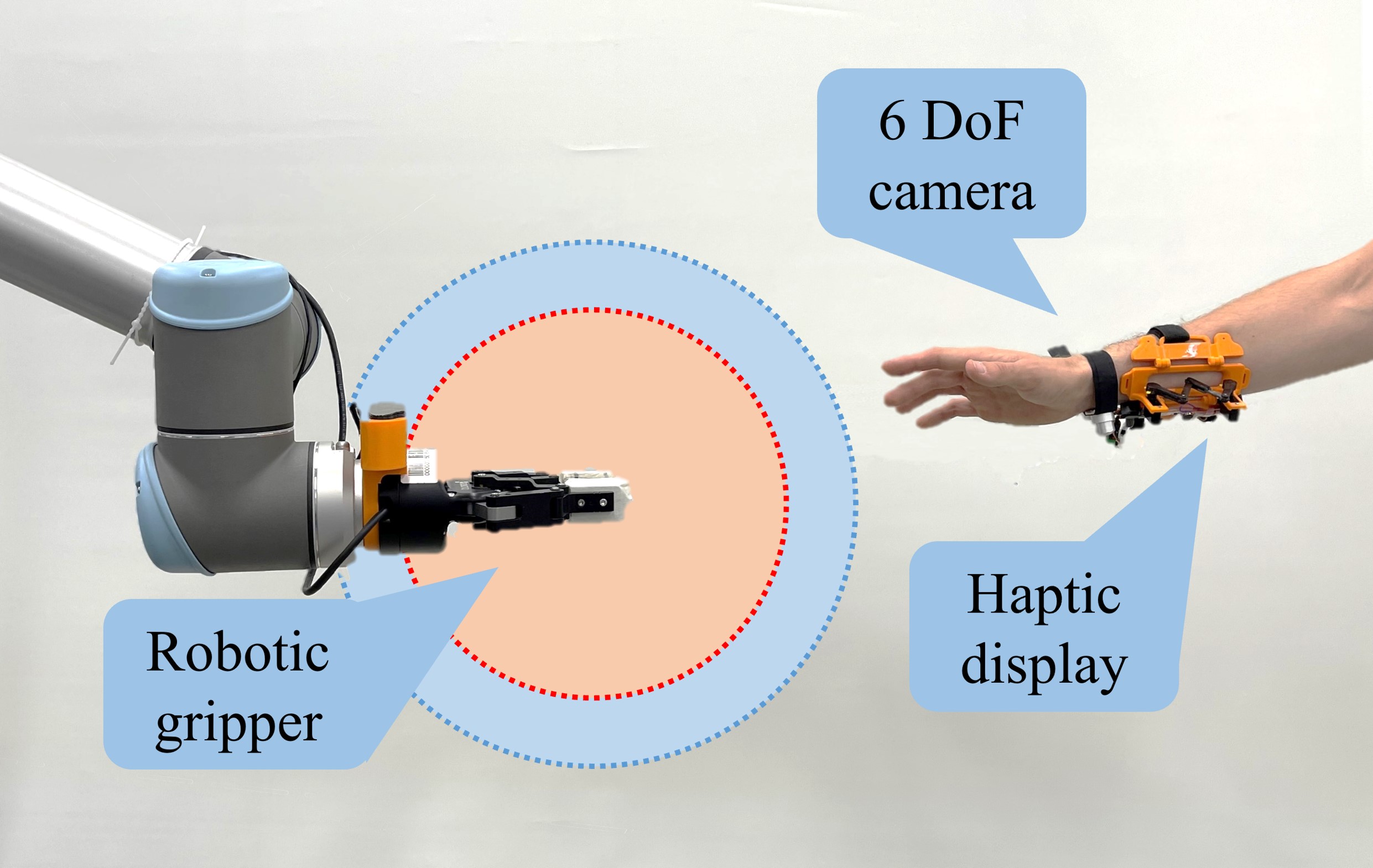}
  \caption{A wearable haptic device applies tactile stimuli when the user intersects the dynamic Haptic Potential Field (blue). The internal sphere (light red) represents volume defined by protective separation distance. }
  \label{fig:mainIdea}
\end{figure}

The first comprehensive framework about principles and requirements to ensure safe Human-Robot Collaboration (HRC) was described in the Technical Specification 15066 by the International Organization for Standardization (ISO/TS 15066) \cite{ISO2016}. Speed and Separation Monitoring (SSM) is considered one of the four safe collaborative scenarios. The main criterion to comply with the SSM scenario is to maintain the protective separation distance (PSD) between robot and human. There are various studies exploring possibilities to mitigate the risk of accidental injury within the scope of SSM. Lasota et al. \cite{safetyRob1},  and Zanchettin et al \cite{safetyRob2} considered the continuous scaling of robot speed based on various distance criteria \cite{SSM}.  Collision avoidance approaches consider online modification of robot trajectory in correspondence with operator position. Mainprice  \cite{RobS3}, and Pereira \cite{Pereira2018} developed models for online estimation of operator's workspace occupancy based on preliminary recorded data of human motion. To our best knowledge, the studies related to SSM research, consider safety only from the robot control system side, and the operator plays a passive role having limited awareness of robot planned actions.

On the other side, several researchers explore sensory augmentation to improve HRC. Wearable haptic interfaces are used to alert the operator about dangerous events and support spatial coordination of processes in a collaborative environment  \cite{Ang2015}. Papanastasiou et al. \cite{Papanastasiou2019} implemented a test-bed comprising Augmented Reality glasses, smartwatches, and a robot with tactile safety skin, to support safe and efficient collaborative assembling operations. The glasses have been used to visualize robot constrained zones and provide instructions in assembly processes, while smartwatches have served only as confirmation buttons. Salvietti et al. \cite{SalviettiRing} proposed an extended HRC concept called Bilateral Haptic Collaboration. To demonstrate this concept, the author designed a finger-worn ring interface to control a robotic gripper remotely. The ring provided vibration stimuli about gripper grasp tightness, and at the same time, the user could teleoperate the robotic gripper motion via the device.

The application of haptic devices for safe HRC has been explored to a limited extent. An approach to promote safe HRC with high-payload industrial robots was proposed in \cite{Vogel2016}. The safety system is comprised of a touch-sensitive tactile floor to measure foot position and four projectors, which provide visual feedback to the operator by rendering static safety areas in the floor around the robot. The authors were aimed to make projected zones dynamic to promote SSM criteria. However, the approach cannot be generalized to account for hands position and the operator has to be constantly aware of the changes in the projection zones. This visual distraction complicates some close HRC scenarios. In \cite{Che2018}, the researchers applied a forearm-worn vibratory bracelet to help the operator avoid collisions with interfering mobile robots. The device transmitted binary information: whether the robot interfered with the human path or not. The cross-condition comparison (visual only vs. visual + haptic) has shown that the additional haptic communication related to the robot intent has significantly improved the user decision-making. 

This paper presents a novel approach for safe HRC to extend SSM safety requirements from the human worker side. We introduce the concept of Haptic Potential Field, where tactile stimuli inform the operator about the dangerous proximity with robots. The HPF depends on the operator's reaction time to the hazardous event and velocities of the end-effector and operator. We propose a methodology for reaction time measurement, which includes the delays caused by software and hardware of the system. The proposed approach is evaluated with a collaborative assembly case. The developed system consists of a wearable haptic display $RecyGlide$ \cite{10.1145/3359996.3364759}, a 6 Degrees of Freedom (DOF) wearable optical mocap by AntiLatency, and a collaborative robot UR10.

The paper is distributed in the following way: Section II describes the proposed approach. In section III the system architecture used for experimental studies is described. Section IV presents a methodology for measuring operators' reaction time and describes an experiment. Section V proposes a collaborative assembly case study to evaluate the proposed approach. Finally, Section VI discusses the conclusions and future work.

\section{Haptic Potential Fields}

The industrial robot has to stop when the distance between the robot and operator becomes less than the protective separation distance to comply with the SSM safety condition \cite{ISO2016}. According to the standard \cite{ISO2016}, there are several ways to define the PSD. In this paper, we measure PSD as a distance between the human hand, provided by the wearable tracker, and the robot Tool Central Point (TCP).

We propose enhancing the safety condition by introducing the concept of Haptic Potential Field. The HPF represents a volume in space, inside of which the operator is haptically informed about possibility of collision with a robotic agent   (Fig. \ref{fig:mainIdea}). The HPF aim is to keep away the human worker from the restrictive volume (RV) or zone, which may be defined by PSD. 
The device has to be suited for manual operations and guarantees substantial stimuli pattern recognition. The selection of particular haptic stimuli patterns is out of the scope of this study.  

The volume occupied by HPF is the Minkowski sum of a convex hull of a set of points $P$, representing Restrictive Volume (RV), and a sphere of radius $r_h$, denoted by B($r_h$) \cite{sweptVol}: 
\begin{equation} \label{conv:1}
 H(r,P)={\rm conv} (P) \oplus B(r_h)
\end{equation}
The $r_h$ depends on the operator's reaction time subject to haptic stimuli $t_{r}$ and velocities of the robot and the operator:

\begin{equation} \label{rh:1}
r_h = 
\begin{cases} 
0, & \mbox{if } \mbox{ $V_{a} < 0$} \\ 
V_{a} \cdot t_{r}, & \mbox{if }\mbox{$V_{a} > 0,$} 
\end{cases}
\end{equation}

where $V_{a}$ is the approaching speed of TCP to operator. We define $t_{r}$ as the average reaction time of the operator, which is measured experimentally as proposed in section IV. 

\begin{figure}[h!]
  \vspace{0.5cm}
  \centering
  \includegraphics[width=0.9\linewidth]{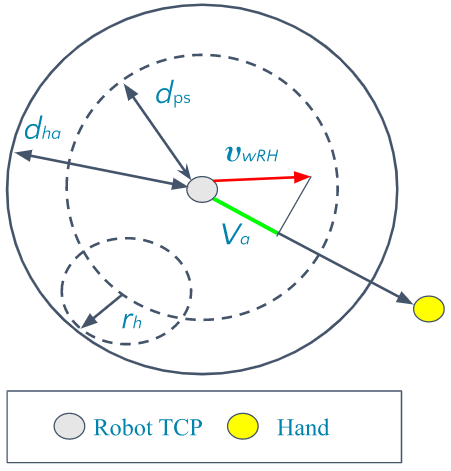}
  \caption{The positive value of approaching speed $V_{a}$ extends $r_h$ and, consequently, increases the radius $d_{ha}$ of the Haptic Potential Field.}
  \label{fig:Formulas}
\end{figure}   

In this study, the RV and HPF are represented by TCP-centered spheres. The radius of RV is equal to PSD and denoted as $d_{ps}$, and the radius of HPF is equal to the Haptic Activation Distance (HAD) $d_{ha}$:  

\begin{equation} \label{had:1}
d_{ha} = min(d_{hmax} , d_{ps} + r_h)
\end{equation}

The constant $d_{hmax}$ is the maximum HAD, which reflects the robot reachability. The relations between components, defining HPV, are illustrated in Fig. \ref{fig:Formulas}, for a positive approaching speed value $V_{a}$. 

In this study, we evaluate the weighted relative velocity $\upsilon_{wRH}$ with the following formula:

\begin{equation}\label{eq:vw}
\centering
\ \boldsymbol{\upsilon_{wRH}} = k_{r} \cdot \boldsymbol{{\upsilon_{r}}} - k_{h} \cdot \boldsymbol{{\upsilon_{h}}}
\end{equation}
where $\boldsymbol{{\upsilon_{r}}}$ and $\boldsymbol{{\upsilon_{h}}}$ are velocities of the TCP and hand, respectively, and the empirical coefficients $k_{h}$, $k_{r}$ are used to calibrate the contribution of velocities to the HAP according to the experimental safety conditions.

\section{System Architecture}

We applied the following equipment to demonstrate our approach: (1) wearable optical mocap $AntiLatency$, (2) wearable multi-modal haptic display $RecyGlide$, (3) UR10 cobot $Universal Robots$. 
 
\subsection{Hardware Description}

The 6 Degrees of Freedom (DOF) wireless wearable optical mocap by $AntiLatency$ provides information about the device's position and orientation within the visibility range of floor-mounted active IR markers. When the device is out of the visibility range, the data is estimated from IMU sensors for a small amount of time. The data updates are provided in real-time at a frequency of 2000 $Hz$. 
\begin{figure}[h!]
  \centering
  \includegraphics[width=0.9\linewidth]{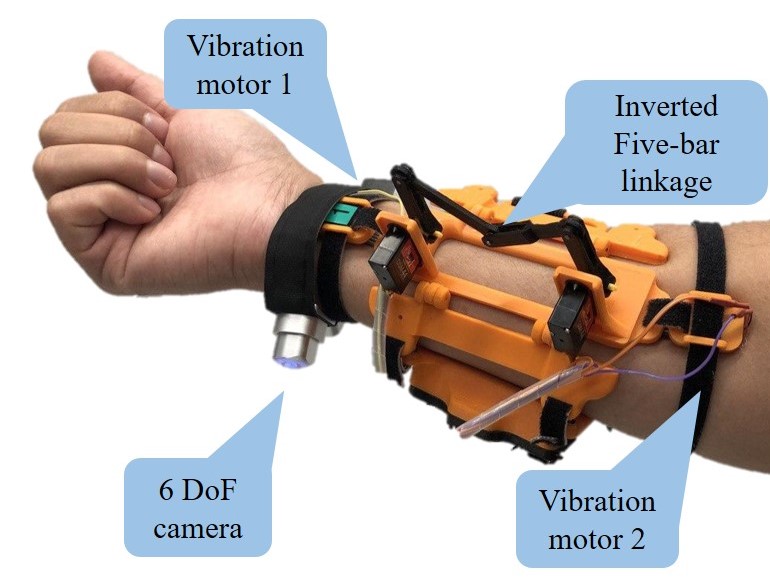}
  \caption{ $RecyGlide$,  a  novel  wearable  haptic  display and AntiLatency  6DoF  motion capturing  camera.}
  \label{fig:Recyglide}
\end{figure}    

$RecyGlide$ is a novel forearm-worn haptic display, which delivers easily recognizable multi-modal stimuli \cite{10.1145/3359996.3364759}. The device provides the sense of touch at one contact point, using an inverse five-bar linkage mechanism, and vibration feedback in two forearm points using two coin vibration motors. Moreover, the $RecyGlide$ has been designed to adapt ergonomically to the user, allowing dexterous movements of the hand for object manipulation. The composition of $RecyGlide$ and mocap on a hand, used in our experimental setting, is shown in Fig. \ref{fig:Recyglide}. 

The collaborative robot used is a UR10 from Universal Robots with a maximum reach radius of 1300 $mm$, and a maximum payload of 10 $kg$. The system's software runs on a PC Intel Core i7-10750H CPU clocked at 2.6GHz  with six cores and 12 threads with 31.8 GB of RAM.  

\subsection{Software Description}
We implemented the software of the system using Python language in the Jupyter Notebook IDE. The Python library urx \cite{URX} was used to control the UR10 program loop. Additionally, we used C\# language and Unity engine to collect position, orientation, and velocity data of the $AntiLatency$ mocap and submit it to the main script via socket connection. The wearable haptic display $RecyGlide$ was controlled from the main script via serial Bluetooth communication. Data collection from the camera, haptic device, robot control, and event handler of human-robot interaction were executed as parallel threads.

\begin{figure}
  \centering
  \vspace{0.5cm}
  \includegraphics[width=1\linewidth]{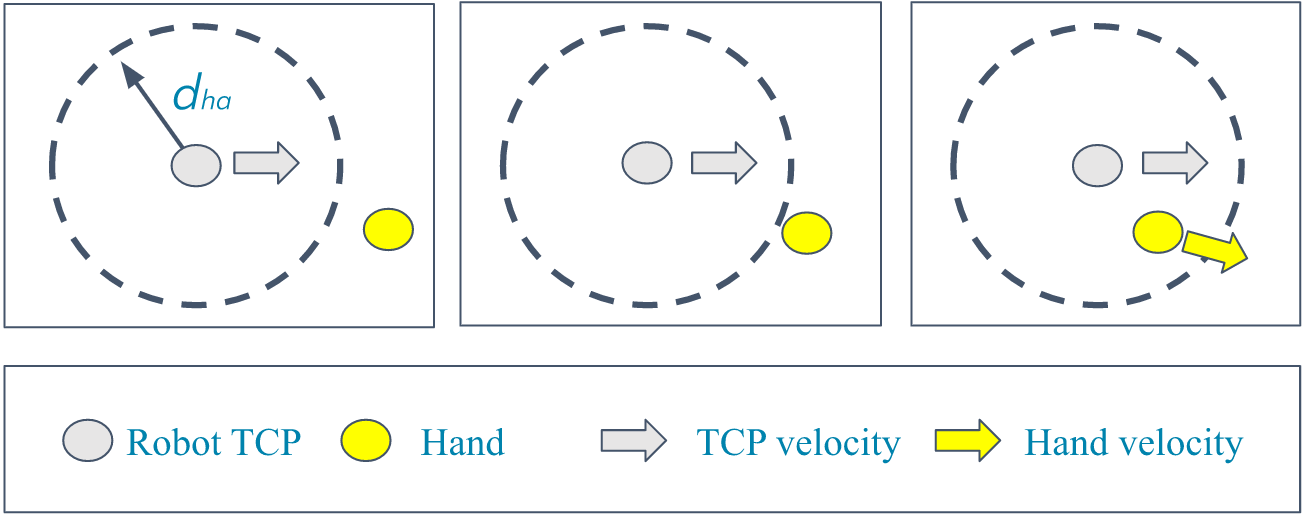}
  \caption{Reaction time measurement per trial. The initial position of hand and end-effector (left), the control script captures HPF interference event and the stopwatch starts (middle), the control script recognizes intentional hand movements event and stopwatch stops (right). }
  \label{fig:RTmeasure}
\end{figure}

\section{Experiment: Operator Reaction Time Measurement}

The first experiment was aimed to identify the user's reaction time subject to applied haptic stimuli when the approaching robot intersected the HPF of constant dimensions. 

Ten right-handed participants (3 females) volunteering to complete the evaluation. The participants were informed about the experiment and agreed to the consent form.

The users were asked to stand in front of the UR10 and to wear the $RecyGlide$ haptic display and the Antilatency 6DoF camera on the right forearm. In addition, the users were asked to close their eyes and to use headphones with white noise to avoid visual and auditory feedback. The haptic feedback stimulation was rendered by the inverted five-bar linkage mechanism of the device at the forearm of the users. Before the experiment, a training session was conducted where the haptic device was adjusted to the participants forearm.

During the experiment, the participants were instructed to move the hand away when they felt the stimulation. In this experiment, we set a minimal hand speed threshold equal to 0.1 $m/s$ to distinguish between intentional hand movement, subject to stimuli, and random hand fluctuations. Moreover, PSD and HAD were equal to 20 $cm$ and 40 $cm$ correspondingly.

\subsection{Reaction Time Measurement}
In each trial, the experimenter recorded the current hand position of the participant referenced to the robot coordinate system. Then, the robot was moving slowly towards the recorded coordinates. The control program loop constantly evaluated the distance between the robot TCP and the user's hand tracked by the wearable mocap device. When the distance became less than HAD, the $RecyGlide$ received a signal to apply stimuli, and we started a stopwatch to measure the reaction time. If the distance became less than PSD, the robot received a signal to stop. When the intentional hand movement was detected, we stopped the stopwatch and obtained the participant's reaction time per conducted trial. In total, 10 measurements per participant were performed. According to this procedure, the delays of hardware and software directly influence the reaction time value. The procedure is illustrated in Fig. \ref{fig:RTmeasure}.

\subsection{Results and Discussion}

\begin{figure}[h!]
 \vspace{0.3cm}
  \centering
  \includegraphics[width=1\linewidth]{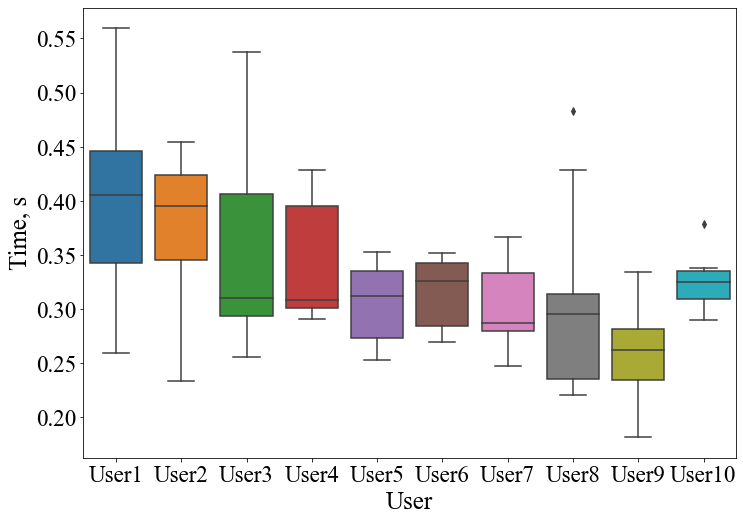}
  \caption{Reaction time of the users to the haptic feedback display.}
  \label{fig:ReactionTime}
\end{figure}

\begin{figure}[h!]
  \centering
  \includegraphics[width=1\linewidth]{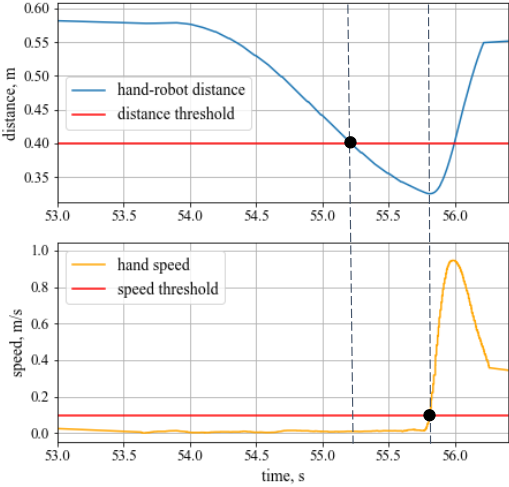}
  \caption{Distance between the user's hand and the End-effector (top), and speed of the user (bottom) per time.}
  \label{fig:ReactionTimeWorst}
\end{figure}

\begin{figure}[h!]
  \vspace{0.5cm}
  \centering
  \includegraphics[width=1\linewidth]{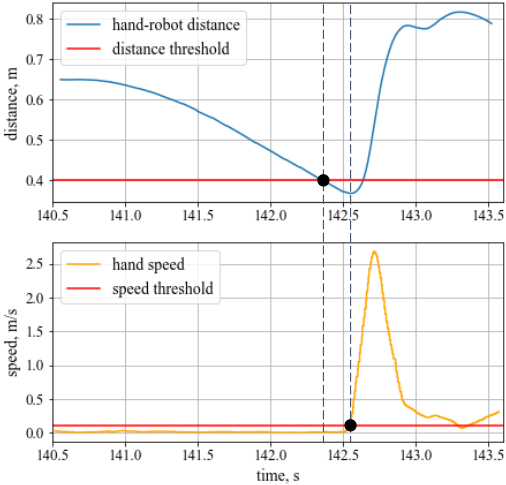}
  \caption{Distance between the user's hand and the end-effector (top), and speed of the user (bottom) per time. }
  \label{fig:ReactionTimeBest}
\end{figure}

The results of the reaction time are summarized in Fig. \ref{fig:ReactionTime}.
The average reaction time is ${0.3243 \pm 0.0715}$ seconds. 
The fastest reaction time of the experiments is 0.1815 seconds, which corresponds to the User 9 (Fig. \ref{fig:ReactionTimeBest}),
and the slowest is 0.5592 seconds, which corresponds to the User 1 (Fig. \ref{fig:ReactionTimeWorst}). The blue line represents the hand-robot distance, and the orange line represents the hand speed. We can observe that the users start to move after the distance threshold is crossed and the haptic stimuli are sent to the hand. 

\section{Collaborative Gearbox Assembly in a Shared Workspace}
This study was aimed to evaluate the influence of haptic feedback on maintaining protective separation distance and the time spent inside the potentially dangerous proximity to the robot. 
\subsection{Experimental setup}

\begin{figure}[h!]
  \vspace{0.5cm}
  \centering
  \includegraphics[width=1\linewidth]{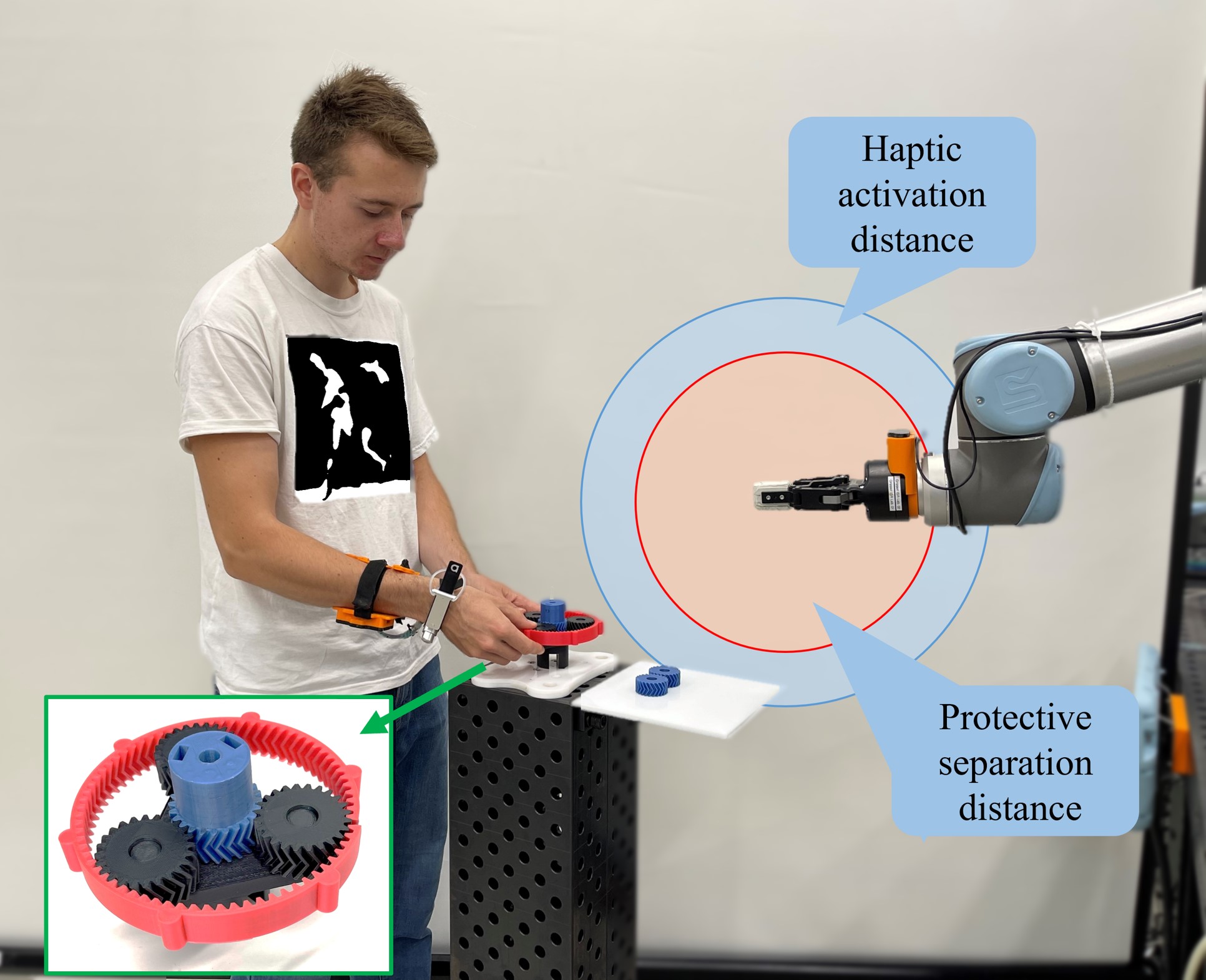}
  \caption{Experiment 2 overview: A participant and a cobot perform collaborative assembly of a herringbone planetary transmission. }
  \label{fig:Experiment 2 overview}
\end{figure}

The second experiment was conducted based on comparative assessment. As in the first experiment, the hand of the participant was equipped with the $Recyglide$ haptic display and Antilatency 6DoF mocap, as shown in Fig. (\ref{fig:Experiment 2 overview}). 
The autonomous assembly of herringbone is challenging and not economical. Therefore, we considered a close collaborative assembly scenario in which the cobot supplied parts to a human worker for manual assembly.
For this reason, the trajectories of the end-effector and human hands were often close to each other.

The participants made several attempts corresponding to the two experimental conditions: 

1 - Visual feedback (V): The user removes the hand from the approaching robot based on the visual channel without haptic feedback.

2 - Visual and haptic feedback (VH): The haptic device informs the user about the dangerous proximity and movements of the robot inside of the dynamic HPF. Thus, the user can be concentrated on the activities relying on haptic feedback.

We set the protective separation distance equal to $0.25$ m. Besides that, the haptic device $Recyglide$ applied the stimuli when the distance between hand and TCP became less than Haptic Activation Distance. HAD is calculated according to the formula \ref{had:1}.
Additionally, the participants were instructed to avoid close proximity to the robot, which is equal to $0.4$ m and denoted as Potentially Dangerous Distance (PDD).  

\subsection{Participants and Procedure}

Five right-handed participants (2 females, aged between 21 and 30 years old) volunteering to complete the evaluation. The participants were informed about the experiment setting and agreed to the consent form. Before the experiment, a training session was conducted. The participants practiced the herringbone gear assembling until they expressed that they were ready. During the experiment, the UR10 robot put the gear parts on the assembler's workbench in the order of assembly. The experiment ended when the herringbone gear was assembled.

\subsection{Results and Discussion}
\begin{figure}[h!]
  \vspace{0.2cm}
  \centering
  \includegraphics[width=1\linewidth]{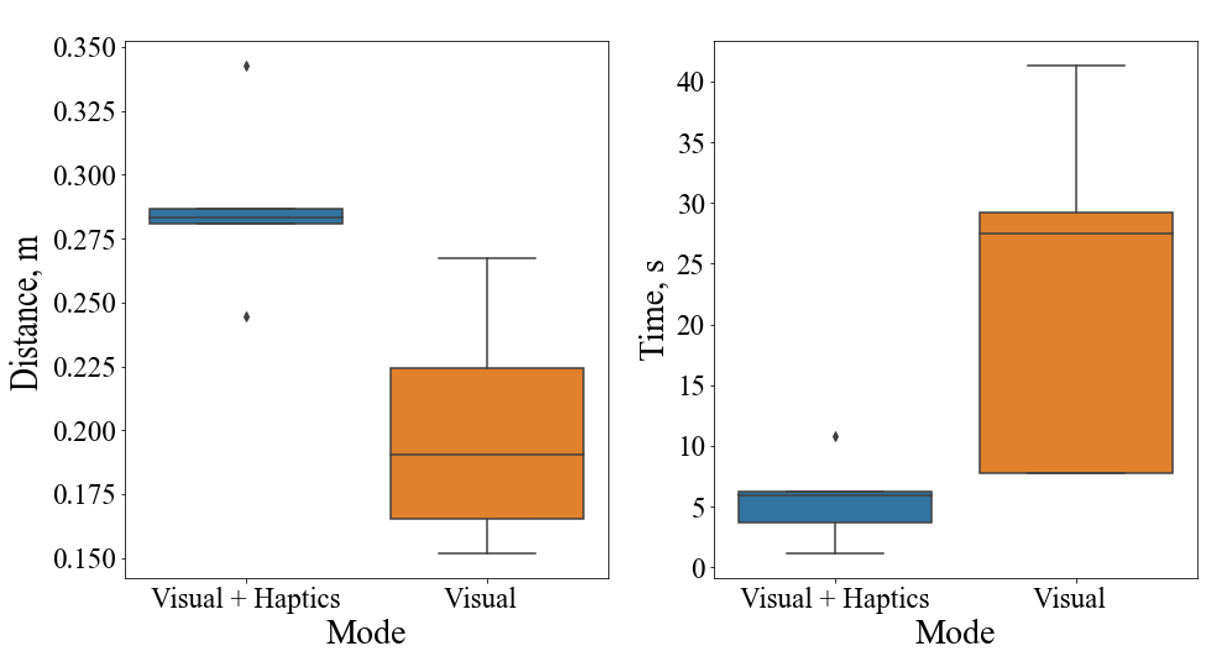}
  \caption{Distance between the cobot TCP and user's hand position (left), and Time (right) for two feedback types (VH and V).}
  \label{fig:timeanddistance}
\end{figure}

We use potentially dangerous distance (PDD) as a simple metric to measure the effect of haptic feedback on the ability of human worker to maintain PSD. The time that the users stayed inside the PDD is presented in Fig. \ref{fig:timeanddistance}. The average time inside the potentially dangerous distance with only visual feedback (V) is 22.69 seconds, and with haptic and visual feedback (VH) is 5.57 seconds. 

In order to evaluate the statistically significant differences between the time with the two conditions (V and VH), the results were analysed using single-factor repeated measures ANOVA, with a chosen significance level of $\alpha<0.05$. According to the ANOVA results, there is a statistically significant difference in the time for each feedback modality, $F(1,8) = 6.4553, p = 0.0346$. The ANOVA showed that the feedback modality implemented (V or VH) significantly influences the time the user stayed inside the potentially dangerous distance between the robot and the user. The average minimum distance between the user and the robot TCP with only visual feedback (V) is 0.1997 m, and with haptic and visual feedback (VH) is 0.2877 m. 

To evaluate the statistically significant differences between the minimum distance with the two feedback modalities, the results were analysed using single-factor repeated measures ANOVA, with a chosen significance level of $\alpha<0.05$. According to the ANOVA results, there is a statistically significant difference in the minimum distance for each of the two feedback modalities, $F(1,8) = 11.3057, p = 9.89\cdot10^{-3}$. The ANOVA showed that the feedback condition implemented significantly influences the minimum distance between the user and the robot TCP.

\begin{figure}[h!]
  \centering
  \includegraphics[width=0.85\linewidth]{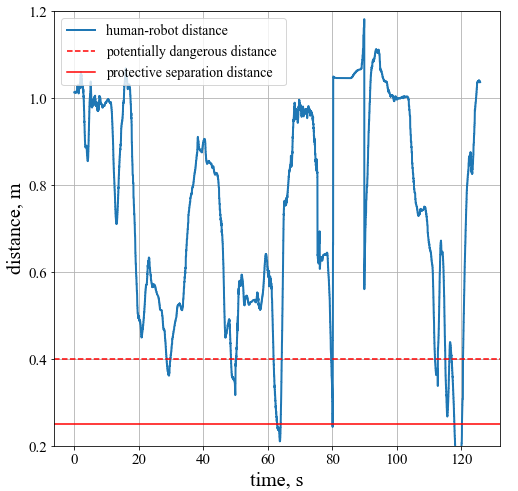}
  \caption{Relative distance between a user's hand and the robot TCP with only visual feedback.}
  \label{fig:distanceV}
\end{figure}

\begin{figure}[h!]
  \centering
  \includegraphics[width=0.85\linewidth]{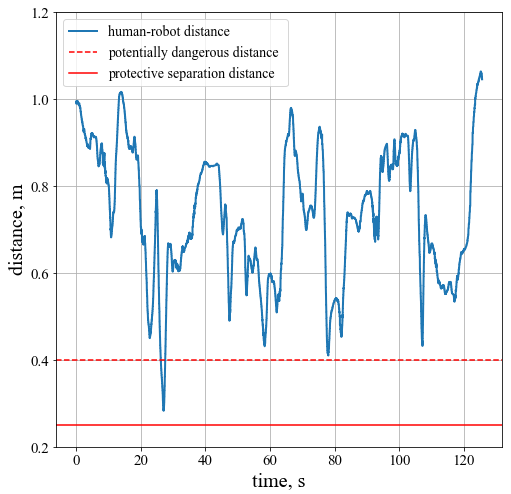}
  \caption{Relative distance between a user's hand and the robot TCP is shown with haptic and visual feedback.}
  \label{fig:distanceVH}
\end{figure}

Fig. \ref{fig:distanceV} shows the relative distance between the user's hand and the robot's TCP when the user perceives only his visual feedback (V). In Fig. \ref{fig:distanceVH} the relative distance between the user's hand and the robot TCP is shown. In this case, the user has access to haptic and visual feedback (VH). The dashed red line represents the PDD (0.4 m). We can observe that crossing the potentially dangerous distance was reduced considerably when the haptic feedback was implemented, and the protective separation distance (0.25 m) was never crossed.

\section{Conclusion and Future Work}

In this study, we proposed a novel approach to promote SSM safety criteria from the human worker side using cheap high-frequency optical 6DoF mocap with IMU sensors and the forearm-worn haptic device $RecyGlide$. 

We introduced the Haptic Potential Field, a field in space, where an operator is informed about dangerous proximity and approaching movements of the cobot by the haptic stimuli. The haptic activation distance is evaluated in correspondence with the average reaction time, measured experimentally, and the instantaneous velocities of the operator's hand and cobot's TCP. 
We evaluated the approach in the collaborative task of herringbone planetary gear assembly based on the cross-condition comparison (visual and haptic feedback and only visual). We found out that our technique improves the average minimum distance between the TCP and operator's hand by 44\% and decrease the time operator spent in dangerous proximity to the robot by 407\%. 

The proposed technology could potentially improve the human-robot interaction and decrease the risk of human harm for collaborative scenarios with industrial robots and cobots. We believe that people with hearing impairment could be especially 
interested in this technology.

For future work, we consider extending the haptic feedback for both hands and explore extended stimuli patterns to suggest to the operator the direction of robot movements and proximity to the protective separation distance. We aim to improve safety metrics by implementing whole-body tracking and account for all links of the robot.  

\section*{Acknowledgment}
The reported study was funded by RFBR and CNRS according to the research project No. 21-58-15006.

\addtolength{\textheight}{-12cm}   

\bibliographystyle{IEEEtran}

\end{document}